\pdfoutput=1

\documentclass[11pt]{article}

\usepackage[final]{acl}

\usepackage{times}
\usepackage{latexsym}

\usepackage[T1]{fontenc}

\usepackage[utf8]{inputenc}

\usepackage{microtype}

\usepackage{inconsolata}

\usepackage{graphicx}

\usepackage{url}            
\usepackage{booktabs}       
\usepackage{multirow}
\usepackage{amsfonts}       
\usepackage{nicefrac}       
\usepackage{microtype}      
\usepackage{xcolor}         
\usepackage{natbib}
\usepackage{bbm}
\usepackage{amsmath}
\usepackage[]{mdframed}
\usepackage{float}
\usepackage{subfigure}
\usepackage{array}
\usepackage[most]{tcolorbox}
\usepackage{listings}

\title{Generating Diverse Hypotheses for Inductive Reasoning}

\author{%
Kang-il Lee$^1$ \quad Hyukhun Koh$^2$ \quad Dongryeol Lee$^1$ \\
\textbf{Seunghyun Yoon$^3$} \quad \textbf{Minsung Kim}$^1$ \quad \textbf{Kyomin Jung}$^{1,2}\thanks{Corresponding authors.}$\\
$^1$Dept. of ECE, Seoul National University \quad $^2$IPAI, Seoul National University \\ $^3$Adobe Research \\
\texttt{\{4bkang,hyukhunkoh-ai,drl123,kms0805,kjung\}@snu.ac.kr}\\
\texttt{syoon@adobe.com}
}

\begin{document}
\maketitle
\begin{abstract}
Inductive reasoning — the process of inferring general rules from a small number of observations — is a fundamental aspect of human intelligence. 
Recent works suggest that large language models (LLMs) can engage in inductive reasoning by sampling multiple hypotheses about the rules and selecting the one that best explains the observations. 
However, due to the IID sampling, semantically redundant hypotheses are frequently generated, leading to significant wastage of compute.
In this paper, we 1) demonstrate that increasing the temperature to enhance the diversity is limited due to text degeneration issue, and 2) propose a novel method to improve the diversity while maintaining text quality.
We first analyze the effect of increasing the temperature parameter, which is regarded as the LLM's diversity control, on IID hypotheses. 
Our analysis shows that as temperature rises, diversity and accuracy of hypotheses increase up to a certain point, but this trend saturates due to text degeneration. 
To generate hypotheses that are more semantically diverse and of higher quality, we propose a novel approach inspired by human inductive reasoning, which we call Mixture of Concepts (MoC). 
When applied to several inductive reasoning benchmarks, MoC demonstrated significant performance improvements compared to standard IID sampling and other approaches.


\end{abstract}

\section{Introduction}

\begin{figure}[t]
    \centering
    \includegraphics[width=\columnwidth]{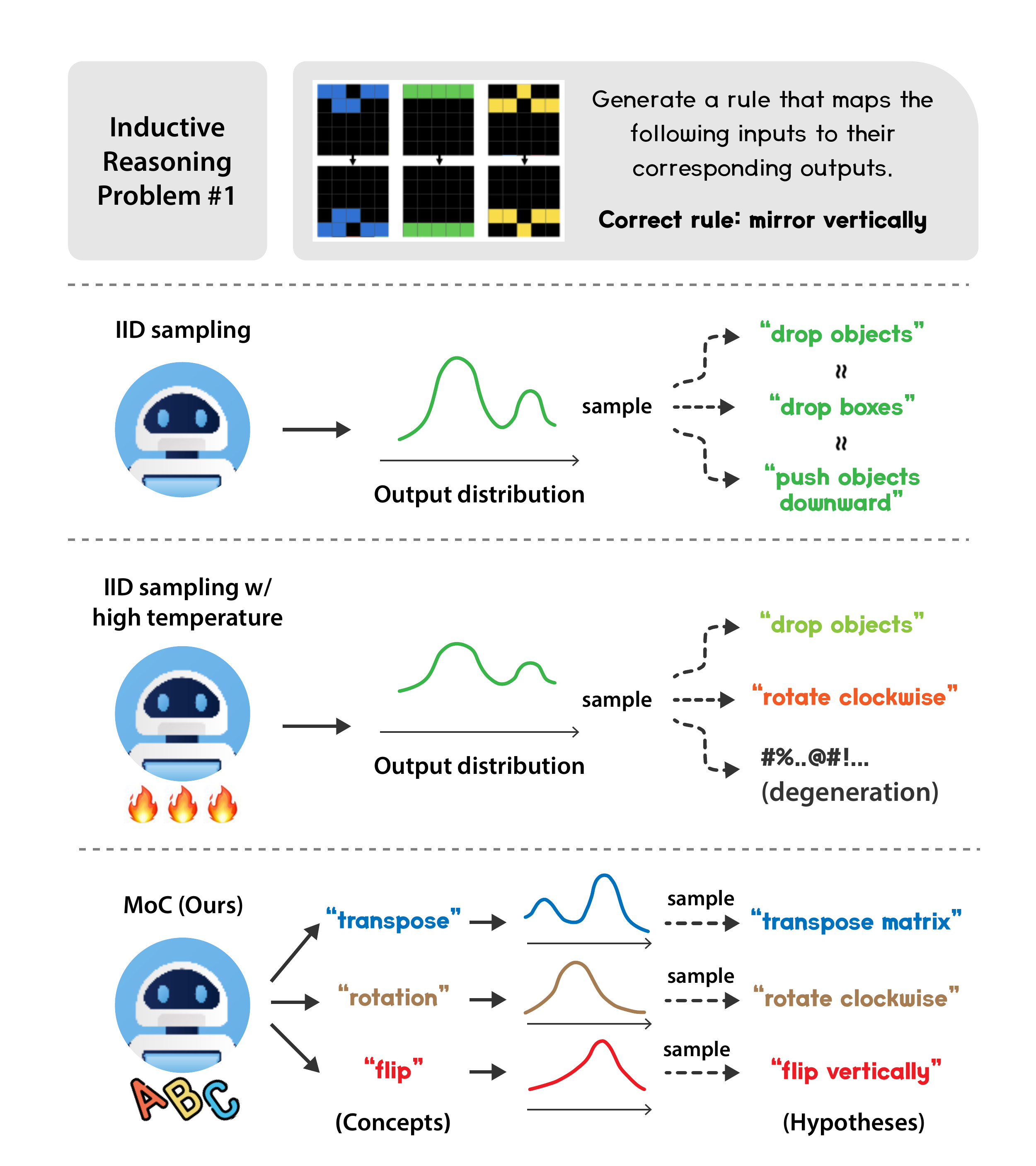}
    \caption{
    A motivation for MoC approach.
    IID sampling frequently generates redundant hypotheses (top). 
    Increasing the temperature leads to frequent occurrences of text degeneration (middle). 
    MoC allows for the generation of diverse hypotheses without a decline in hypothesis quality (bottom).
    }
    \label{figure_moc}
\vspace{-12pt}
\end{figure}

\textit{Inductive reasoning}, inferring general rules that explain a small number of observations (Figure \ref{figure_moc}), is a key factor of human intelligence \citep{humanlevel, chollet2019measure}. 
Recently, there has been exploration into automatic inductive reasoning using large language models (LLMs). 
Trained with vast corpora and instruction tuning, LLMs can function as zero-shot rule proposers when given an inductive reasoning problem \citep{phenomenal}. 
By leveraging this capability, a new paradigm in automatic inductive reasoning has emerged: feeding observations (often referred to as train examples) as prompts into the LLM, sampling multiple hypotheses about the rule, and selecting the one that best explains the observations \citep{hypothesis_search, greenblatt, brown2024largelanguagemonkeysscaling}.
However, when the hypotheses are generated by LLMs are fundamentally IID, it often leads to redundant sampling of semantically identical hypotheses. 
Such redundancy reduces the diversity of hypotheses and results in a significant waste of compute.



To address aforementioned problem, we begin by increasing the temperature, which is often considered a diversity parameter for LMs, in IID sampling setup.
Our findings indicate that while both diversity and accuracy improve up to a certain point with increased temperature, they quickly reach saturation. 
This is because raising the temperature deteriorates the quality of hypotheses, and increases the rate of text degeneration \citep{topp}.

To generate diverse hypotheses while maintaining their quality, we propose a simple yet effective method called Mixture of Concepts (MoC). 
Inspired by human inductive reasoning \citep{geb, mitchell2021abstraction}, our approach consists of two stages: \textit{concept proposal} and \textit{hypothesis generation}. 
In the concept proposal, we instruct the LLM to generate ``a list of $K$ concepts'' that are likely to aid in the formation of hypotheses, and semantically non-redundant concepts are generated \textit{sequentially}. 
In the hypothesis generation, these concepts are parsed and each concept is provided as a hint for generating hypotheses.
Our approach allows for the \textit{parallel} generation of semantically diverse hypotheses without hurting the quality of hypotheses (Figure \ref{figure_moc}).

We conduct experiments on four inductive reasoning datasets from distinct domains.
Our methodology significantly improves performance across four LLMs by generating diverse hypotheses while maintaining their quality. 
Compared to vanilla IID sampling with the same number of generated hypotheses, MoC boosts average accuracy by 4.5\%p for GPT-4o-mini and 5.0\%p for Llama-3.1-70B-Instruct as base LLM, respectively.
Our analysis also shows that MoC enables the LLMs to crack challenging problems that are impossible to solve through IID sampling within a practical compute budget.

Our contributions are as follows:

\begin{itemize}
    \item We analyze how the diversity and accuracy of IID hypotheses generated from LLMs vary with a change in temperature.
    \item We propose Mixture of Concepts (MoC), a methodology enabling more diverse and parallel hypothesis generation in LLMs.
    \item Our experimental results on four datasets and four models show that MoC significantly enhances both the efficiency and performance of inductive reasoning capability of LLMs.
\end{itemize}

\section{Hypothesis Diversity and Temperature}
In this section, we provide a description of the problem statement and the baseline method. Subsequently, we analyze the impact of temperature on the diversity and accuracy of hypotheses generated by an LLM.

\subsection{Problem Statement and Baseline}
\label{problem_statement_and_baseline}
\paragraph{Problem Statement}
We consider inductive reasoning problem with $n$ train examples $D_{train}=\{(x_1, y_1), (x_2, y_2), ..., (x_n, y_n)\}$ and $m$ test examples $D_{test}=\{(x_1', y_1'), (x_2', y_2'), ..., (x_m', y_m')\}$.
There is a function $f$ that maps all the inputs to corresponding outputs: $f(x_i)=y_i$ and $f(x_j')=y_j'$ for all $i\in[n]$ and $j\in[m]$, where $[n]=\{1, 2, ..., n\}$.
The goal of this task is to predict $\{y_1', y_2', ..., y_m'\}$, given the $D_{train}$ and $\{x_1', x_2', ..., x_m'\}$.

\paragraph{Simple Baseline}
Here, we describe a simple baseline method that we use in our experiments.
We adopt the setup defined in recent inductive reasoning and Programming-by-Example (PBE) literature, where the function $\hat{f}$ is first inferred and the test outputs are obtained as $\{\hat{f}(x_j')\}_{j\in[m]}$.
The function $\hat{f}$ is implemented as a Python function, unlike traditional PBE works using domain-specific lanuguages (DSLs).

We prompt LLMs with instruction to generate a hypothesis about $f$ in natural language form and then implement it in Python, following the recent LLM inductive reasoning works \citep{phenomenal, hypothesis_search}.
We sample a fixed number of $K$ responses and extract the Python function from these responses, forming a hypothesis pool $\{\hat{f}_1, \hat{f}_2, ..., \hat{f}_K\}$.
Among these hypotheses, the one that perfectly explains the train examples, i.e. $y_i=\hat{f}_k(x_i)$ for all $i\in[n]$, is picked and submitted to be validated on the test examples.
If there is no such hypothesis, we regard the problem as not being solved correctly.
We consider that a problem is solved correctly if $\hat{f}_k$ passes every test cases: $y_j'=\hat{f}_k(x_j')$ for all $j\in[m]$.


\paragraph{Duplicate Hypotheses}
The issue with the aforementioned simple baseline is that, since it samples $K$ hypotheses from IID distributions, a significant portion of these hypotheses may be semantically redundant. 
If the hypothesis is correct, this redundancy does not pose a problem; however, if an incorrect hypothesis is sampled multiple times, it results in a serious waste of the generation budget.

To investigate whether this redundancy is problematic, we analyze how many semantically unique hypotheses are among the $K$ hypotheses, for the instances that the baseline failed to solve correctly. 
We implement the baseline using GPT-4o-mini (\texttt{gpt-4o-mini-2024-07-18}) with temperature value of 1 and conduct the experiment in the List Functions dataset \citep{listfn}.
In order to count the number of semantically unique hypotheses, a method to determine whether two hypotheses were semantically identical or different is needed.
Since we represent the hypotheses as Python programs, we consider two hypotheses to be identical if both functions return the same outputs for the same inputs.

\begin{table}[H]
\centering
\small
\begin{tabular}{@{}llllll@{}}
\toprule
$K$          & 2    & 4    & 8    & 16   & 32   \\ \midrule
$N$ & 1.40 & 1.79 & 2.92 & 4.56 & 7.89 \\ \bottomrule
\end{tabular}
\caption{Average number of unique programs ($N$) in $K$ generated hypotheses for failed instances on List Functions dataset.}
\label{table_unique}
\vspace{-4pt}
\end{table}

As shown in Table \ref{table_unique}, we can observe that among the incorrect hypotheses, the majority of hypotheses are semantically redundant, especially when $K$ is large. 
This indicates the need to generate a more diverse set of hypotheses that are semantically distinct for a more efficient hypothesis search.

\subsection{Impact of Temperature on Diversity and Accuracy}
\label{impact_of_temperature}

\paragraph{Experiment on Temperature}
One of the most straightforward ways to increase diversity when sampling from a language model is by raising the temperature parameter. 
In this section, we analyze how temperature affects hypothesis diversity and accuracy in inductive reasoning tasks. 
We focus on the changes that occur when varying the temperature in the baseline method, within the datasets of List Functions and MiniARC \citep{miniarc}.

\begin{figure}[H]
    \centering
    \includegraphics[width=0.7\columnwidth]{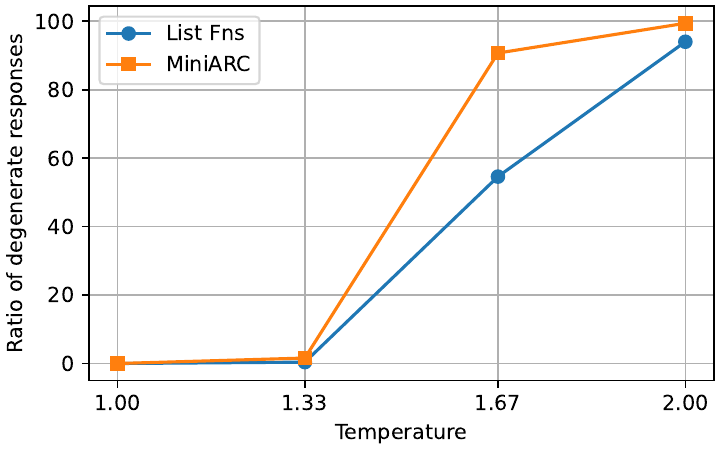}
    \vspace{-5pt}
    \caption{
    Ratio (\%) of degenerate responses.
    }
    \label{figure_degen}
    \vspace{-8pt}
\end{figure}

We conduct experiments with a total of seven temperature settings, ranging from 0 to 2 in increments of 0.33. 
For temperatures of 1.67 and 2.0, there are substantial portion of responses that are degenerated (i.e. the Python function cannot be parsed from the model's response) (Figure \ref{figure_degen}).

Therefore, we use top-$p$ sampling \citep{topp} with $p=0.95$ for these temperature settings, which eliminates degenerate responses entirely.
Additionally, by varying the number of sampled responses $K$, we observe how both $K$ and temperature affect the diversity of hypotheses.

\begin{figure}[H]
\vspace{-18pt}
\begin{center}
\subfigure[List Functions]{%
    \label{list_fn_temperature_div}%
    \includegraphics[width=0.5\columnwidth]{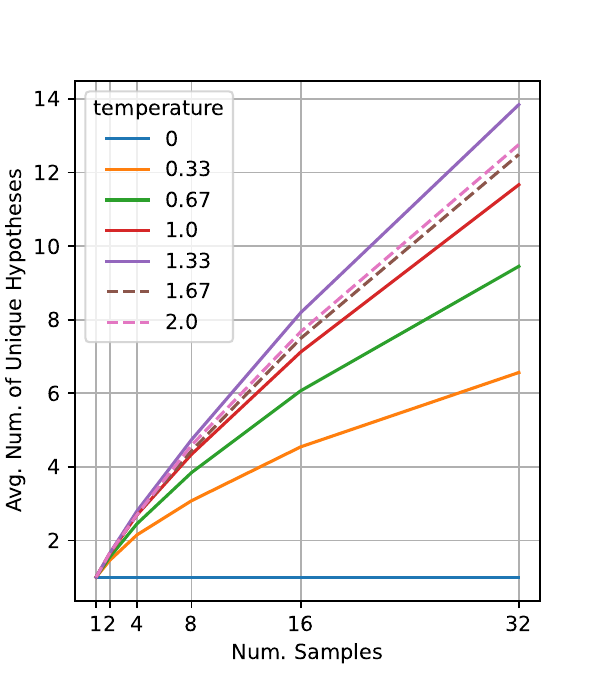}}%
\subfigure[MiniARC]{%
    \label{miniarc_temperature_div}%
    \includegraphics[width=0.5\columnwidth]{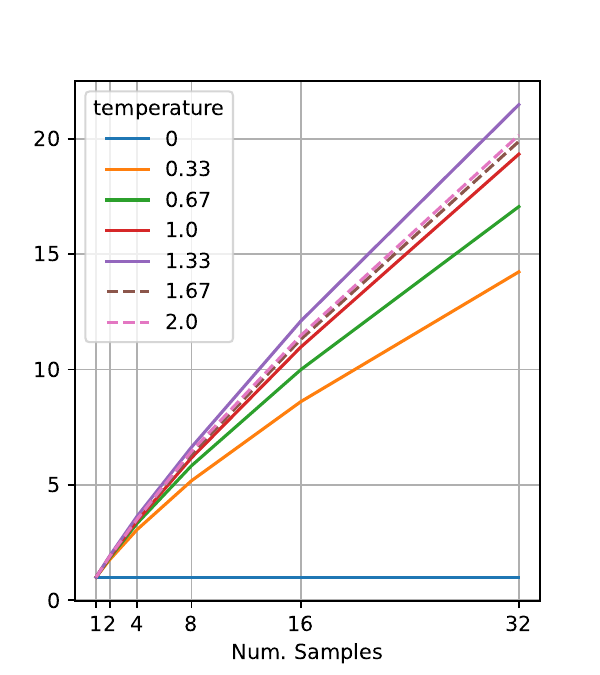}}%
\end{center}
    \vspace{-8pt}
    \caption{GPT-4o-mini hypothesis diversity on two domains. For the temperature 1.67 and 2.0, we used top-$p$ sampling with $p=0.95$. Results are averaged over 5 runs.}
\label{figure_temperature_div}
\end{figure}

In Figure \ref{figure_temperature_div}, it is observed that larger $K$ and higher temperature correspond to larger number of semantically unique programs.
However, beyond the temperature of 1.67, no significant difference compared to 1.33 is observed.
This plateau may be attributed to the side effect of using top-$p$ sampling.
This dilemma, whether to use top-$p$ sampling or not, is the primary reason why higher temperature values do not lead to additional diversity improvements.


\begin{figure}[H]
\vspace{-8pt}
\begin{center}
\subfigure[List Functions]{%
    \label{list_fn_temperature_acc}%
    \includegraphics[width=0.5\columnwidth]{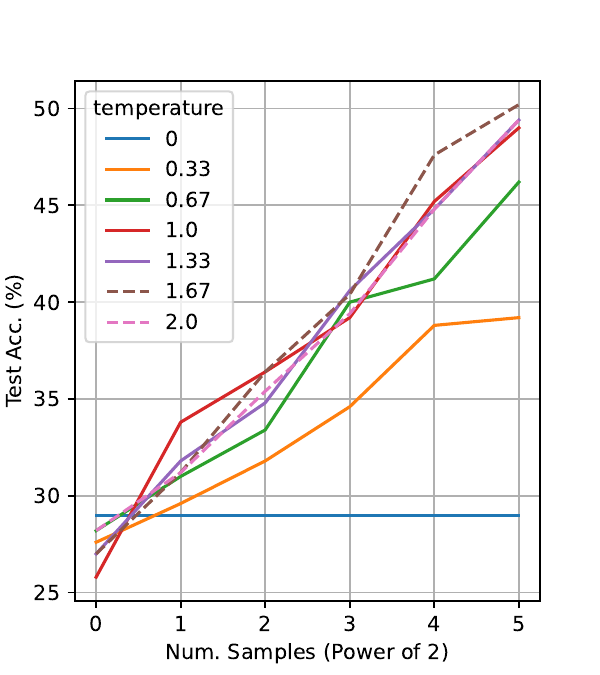}}%
\subfigure[MiniARC]{%
    \label{miniarc_temperature_acc}%
    \includegraphics[width=0.5\columnwidth]{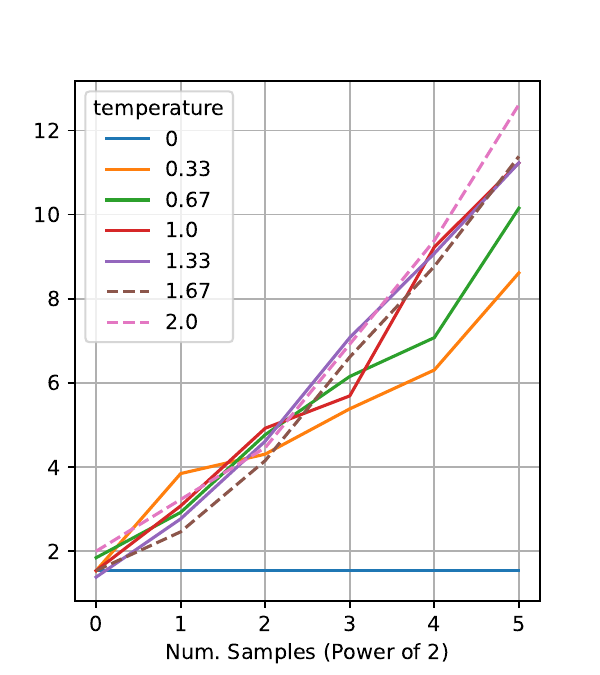}}%
\end{center}
    \vspace{-8pt}
    \caption{GPT-4o-mini performance on two domains. For the temperature 1.67 and 2.0, we used top-$p$ sampling with $p=0.95$. Results are averaged over 5 runs.}
\label{figure_temperature_acc}
\vspace{-4pt}
\end{figure}

\begin{figure*}[t]
    \centering
    \includegraphics[width=0.87\textwidth]{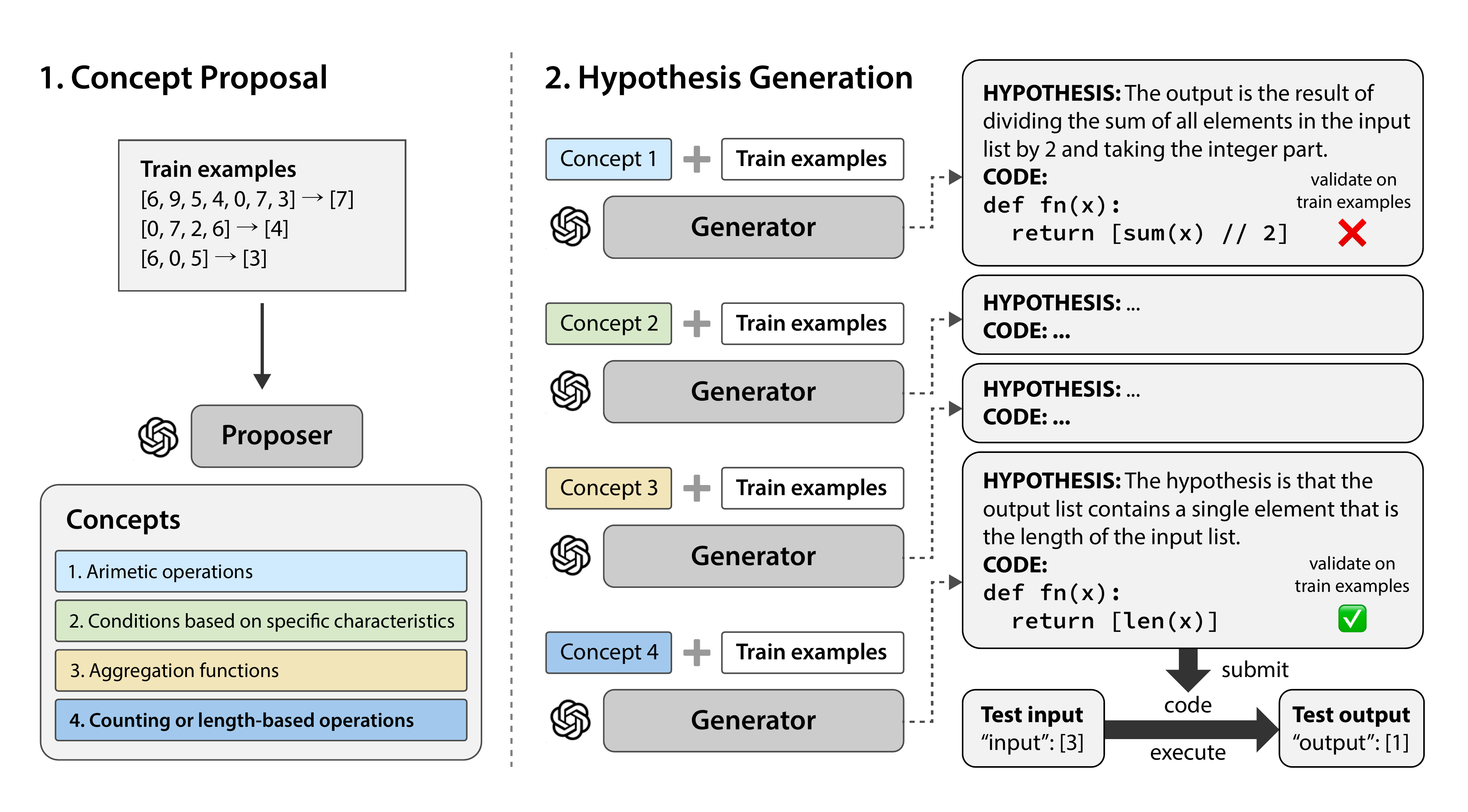}
    \vspace{-10pt}
    \caption{
    An overview of our Mixture of Concepts approach. We generate $K$ distinct concepts (left) and feed them into the LLM separately for hypothesis generation (right).
    }
    \label{figure_structure}
\end{figure*}

As shown in Figure \ref{figure_temperature_acc}, the accuracy shows a similar trend with that of diversity.
Higher temperature values have advantage over lower values, but only up to temperature 1.0.
It suggests that the quality of generated hypotheses starts to decline before text degeneration occurs, making additional hypotheses no longer helpful to the accuracy \citep{peeperkorn2024temperature}.
Therefore, in order to further improve performance, it is necessary to find a reliable method to increase semantic diversity while maintaining the quality of the hypotheses.



\section{Mixture of Concepts}
To achieve aforementioned goal, we propose automatic inductive reasoning framework called Mixture of Concepts (MoC). 

When humans engage in inductive reasoning, they begin by carefully observing examples and considering which \textit{concept} or \textit{template} might be effective in identifying the underlying rule \citep{mitchell2021abstraction}. 
For instance, if presented with integer lists as depicted in Figure \ref{figure_structure}, one might initially contemplate concepts such as ``arithmetic operations.'' 
Using this concept, human would re-examine the given examples to discern what the hidden rule might be. 
However, in this instance, the correct rule is to output the length of the list, and the concept of ``arithmetic operations'' do not aid in discovering this rule. 
In such scenarios, the best strategy is to revisit the given examples from a different point of view and to contemplate novel concepts and rules.
This process can be repeated until the correct rule is identified \citep{geb}.

Inspired by this cognitive process, we propose a framework composed of two stages: \textit{concept proposal} and \textit{hypothesis generation}.

\paragraph{Concept Proposal}
Through empirical investigation, we discovered that when instructing an LLM to generate a list of items, the generated items are rarely semantically redundant. 
This capability can be attributed to the autoregressive nature of the LLM. 
Each item generated by the LLM is used as context for generating the subsequent item, and the LLM tends to produce an item different from those it has already generated.

By leveraging this property, an LLM is instructed to list $K$ elementary concepts that are likely to help the formulation of a hypothesis for given observations.
For instance, given an inductive reasoning problem shown in Figure \ref{figure_structure}, the LLM generates a list of 4 distinct concepts (e.g. Arithmetic operations, Aggregation functions, etc.).
Additionally, we instructed the LLM to generate the concepts in JSON format, ensuring that the concepts can be parsed more reliably in the subsequent stage.

\begin{figure*}[t]
    \centering
    \includegraphics[width=0.85\textwidth]{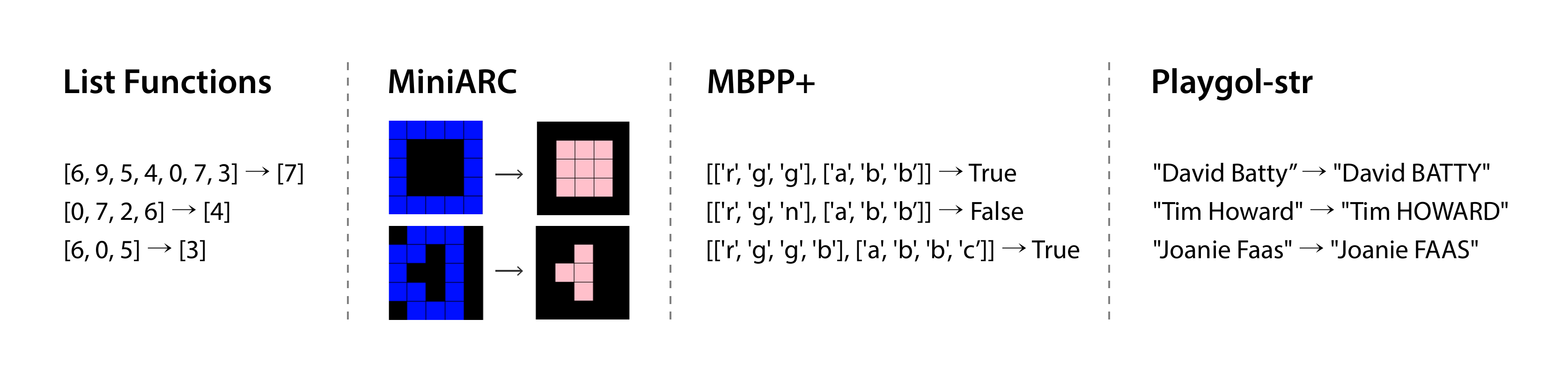}
    \vspace{-10pt}
    \caption[Example problems in each of four datasets we study.
    We graphically display the MiniARC examples to help the reader understand.
    ]{
    Example problems in each of four datasets we study.
    We graphically display the MiniARC examples to help the reader understand.
    }
    \label{figure_domain}
\end{figure*}

\paragraph{Hypothesis Generation}
Next, we parse the concepts and feed each concept to the LLM.
The LLM creates a natural language hypothesis and a Python code implementation based on the given concept.
Here, a concept is provided in the instruction prompt as hint, so that the hypothesis can be conditioned on the given concept.

For instance, the 1st concept in Figure \ref{figure_structure} (Arithmetic operations) conditions the generation to the wrong path and results in incorrect hypothesis. 
Meanwhile, the 4th concept (Counting or length-based operations) guides the generation to semantically distinct direction, with correct output of returning the length of given input list (for full prompt details, see Appendix \ref{appendix_prompt}).

Finally, one of the hypotheses (codes) that satisfies all the given train examples is submitted as the final hypothesis. 
The submitted hypothesis is considered correct if it satisfies all the test examples.



\begin{table*}[t]
\small
\centering
\begin{tabular}{@{}lllllll@{}}
\toprule
\textbf{Model}                    &          & List Fns    & MiniARC       & MBPP+       & Playgol-str   & Average              \\ \midrule
\multirow{3}{*}{GPT-4o-mini}      & Baseline & 43          & 5.4           & 33          & 75.3          & 39.2                 \\
                                  & IHR (\textit{T}=2, \textit{N}=4) & 39            & 5.4           & -             & -             & -                     \\
                                  & MoC      & \textbf{49} & \textbf{8.5}  & \textbf{40} & \textbf{77.3} & \textbf{43.7 (+4.5)} \\ \midrule
\multirow{3}{*}{GPT-4o}           & Baseline & 47          & 9.2           & 53          & 81.3          & 47.6                 \\
                                  & IHR (\textit{T}=2, \textit{N}=4) & 52            & 11.5           & -             & -             & -                     \\
                                  & MoC      & \textbf{53} & \textbf{12.3} & \textbf{55} & \textbf{83.3} & \textbf{50.9 (+3.3)} \\ \midrule
\multirow{2}{*}{Llama 3.1 70B} & Baseline & 40          & \textbf{9.2}  & 45          & 64            & 39.6                 \\
                                  & MoC      & \textbf{45} & 7.7           & \textbf{55} & \textbf{70.7} & \textbf{44.6 (+5.0)} \\ \midrule
\multirow{2}{*}{Qwen2.5 72B}  & Baseline & 46          & 6.9           & 45          & 76.7          & 43.7                 \\
                                  & MoC      & \textbf{53} & \textbf{8.5}  & \textbf{51} & \textbf{78.7} & \textbf{47.8 (+4.1)} \\ \bottomrule
\end{tabular}
\caption{Test accuracy (\%) of IID baseline, Iterative Hypothesis Refinement (IHR) \citep{phenomenal}, and our MoC, with 8 hypotheses generated. The numbers in parentheses indicate the improvement compared to the baseline.}
\label{table_main}
\end{table*}

\section{Experiments}
\subsection{Datasets}

\begin{table*}[t]
\small
\centering
\begin{tabular}{@{}lllllll@{}}
\toprule
\textbf{Model}                             &         & List Fns      & MiniARC       & MBPP+         & Playgol-str   & Average               \\ \midrule
\multirow{2}{*}{GPT-4o-mini}      & Baseline & 4.26          & 6.31          & 3.76          & 2.51          & 4.21                  \\
                                  & MoC      & \textbf{4.75} & \textbf{6.52} & \textbf{4.40} & \textbf{2.63} & \textbf{4.58 (+0.37)} \\ \midrule
\multirow{2}{*}{GPT-4o}           & Baseline & 4.19          & 6.19          & 3.99          & 2.10          & 4.12                  \\
                                  & MoC      & \textbf{5.08} & \textbf{6.80} & \textbf{4.64} & \textbf{2.31} & \textbf{4.71 (+0.59)} \\ \midrule
\multirow{2}{*}{Llama 3.1 70B} & Baseline & 4.81          & 6.35 & 4.10          & 3.07          & 4.58                  \\
                                  & MoC      & \textbf{6.07} & \textbf{7.23} & \textbf{5.35} & \textbf{3.66} & \textbf{5.58 (+1.00)} \\ \midrule
\multirow{2}{*}{Qwen2.5 72B}  & Baseline & 4.22          & 5.72          & 3.05          & 2.24          & 3.81                  \\
                                  & MoC      & \textbf{5.26} & \textbf{6.28} & \textbf{3.82} & \textbf{2.75} & \textbf{4.53 (+0.72)} \\ \bottomrule
\end{tabular}
\caption{Average number of unique hypotheses in 8 hypotheses. The numbers in parentheses indicate the improvement compared to the baseline.}
\label{table_diversity}
\end{table*}

We validate our method on four datasets from distinct domains as shown in Figure \ref{figure_domain}.

\paragraph{List Functions}
The List Functions dataset \citep{listfn} evaluates inductive reasoning ability in inferring list transform functions.
The inputs and outputs are integer list, possibly an empty list.
The transformation function covers broad range of list and arithmetic operators such as indexing, slicing, counting, sorting, etc.
We randomly sample a subset of 100 instances from the 250 instances of the dataset for evaluations in our work.


\paragraph{MiniARC}
MiniARC \citep{miniarc} is a simplified version of Abstraction and Reasoning Corpus (ARC) dataset \citep{chollet2019measure}.
In ARC, the inputs and outputs are 2D grids following specific transformation pattern.
The transformation functions are explicitly grounded on the cognitive priors of humans, such as objectness, goal-directedness, arithmetic and basic geometry \citep{spelke2007core}.
Compared to original ARC, MiniARC's inputs and outputs are always 5x5 grids, greatly reducing the complexity of the problems.
Nonetheless, MiniARC is still extremely challenging for state-of-the-art LLMs \citep{phenomenal}.
Following the previous works, we input the MiniARC grid into the LLMs in the form of a nested list.

\paragraph{MBPP+}
MBPP+ \citep{evalplus} expands a program synthesis dataset MBPP \citep{austin2021program} with 35x more test cases.
It covers basic Python programming tasks written by humans.
The abundance of test cases avails utilizing MBPP+ as an inductive reasoning dataset.
We arrange each instance such that it has 8 train examples and 6 test examples, and randomly sample a subset of 100 instances for evaluation.

\paragraph{Playgol-str}
The string transformation domain is a practical application of automatic inductive reasoning that many users actively utilize \citep{Gulwani, flashfill++, spreadsheetcoder}.
We evaluate our method on the real-world string transformations dataset introduced by \citet{playgol}.
In our work, we refer it using the name of their proposed method, \textit{Playgol-str}.
We randomly sample a subset of 150 instances for evaluation.

\begin{table*}[t]
\small
\centering
\begin{tabular}{@{}lllllll@{}}
\toprule
\textbf{Model}                 &           & List Fns    & MiniARC      & MBPP+       & Playgol-str   & Average       \\ \midrule
\multirow{2}{*}{GPT-4o-mini}   & Greedy    & 29          & \textbf{1.5} & 24          & 52.7          & 26.8          \\
                               & MoC (\textit{K}=1) & \textbf{34} & 0.8          & \textbf{25} & \textbf{63.3} & \textbf{30.8} \\ \midrule
\multirow{2}{*}{GPT-4o}        & Greedy    & 39          & \textbf{4.6} & \textbf{31} & \textbf{69.3} & \textbf{36.0} \\
                               & MoC (\textit{K}=1) & \textbf{42} & 3.8          & 24          & \textbf{69.3} & 34.8          \\ \midrule
\multirow{2}{*}{Llama 3.1 70B} & Greedy    & \textbf{32} & 2.3          & 31          & \textbf{44.7} & 27.5          \\
                               & MoC (\textit{K}=1) & 30          & \textbf{3.1} & \textbf{34} & \textbf{44.7} & \textbf{28.0} \\ \midrule
\multirow{2}{*}{Qwen2.5 72B}   & Greedy    & 30          & 0.8          & 29          & \textbf{58}   & 29.5          \\
                               & MoC (\textit{K}=1) & \textbf{31} & \textbf{3.8} & \textbf{32} & 54.7          & \textbf{30.4} \\ \bottomrule
\end{tabular}
\caption{Test accuracy (\%) of baseline and MoC using greedy decoding. MoC increases the diversity of hypotheses when generating multiple hypotheses, so it offers little benefit in cases where only a single hypothesis is generated using greedy decoding.}
\label{table_greedy}
\end{table*}

\subsection{Main Results}
We compare Mixture of Concepts (MoC) with simple baseline described in section \ref{problem_statement_and_baseline}.
Each method is implemented with two proprietary and two open-source LLMs: GPT-4o-mini (\texttt{gpt-4o-mini-2024-07-18}), GPT-4o (\texttt{gpt-4o-2024-08-06}), Llama-3.1-70B-Instruct, and Qwen2.5-72B-Instruct.
If not otherwise specified, the temperature is set to 1.0 for all experiments, because temperature higher than that does not bring any performance gain as observed in section \ref{impact_of_temperature}.
Also, we set the number of samples $K=8$ here.
Results with $K=4$ and $K=16$ are in Appendix \ref{appendix_additional_results}.

In Table \ref{table_main}, our MoC approach shows significant improvement in overall test accuracy, demonstrating the effectiveness of our approach.
In the MiniARC dataset, all LLMs generally exhibit low performance, which is largely due to the fact that ARC-variants often require visual priors that LLMs are not expected to possess.

\paragraph{Comparison with Hypothesis Refinement}
Iterative Hypothesis Refinement (IHR) \citep{phenomenal} involves validating the generated hypotheses against the training examples, and if none of the hypotheses pass, the hypothesis with highest train accuracy is refined using execution feedback. 
We conduct experiments of IHR in a setting of $T=2$ and $N=4$, similar to the case where $K=8$. 
This means there are 2 iterations, with 4 hypotheses generated per iteration.
Additionally, when translating the generated hypotheses into Python code, we insert the training examples in the input prompt, resulting in a significant performance improvement.

As shown Table \ref{table_main}, iterative hypothesis refinement performs well with GPT-4o, but with weaker GPT-4o-mini, it does not outperform the baseline. 
However, our MoC methodology consistently shows performance improvements regardless of the model's base performance.

\paragraph{Hypothesis Diversity}
To analyze the direct impact of MoC on hypothesis diversity, we observe the number of unique hypotheses under the same conditions as the previous accuracy experiment ($K=8$). 
The number of unique hypotheses is measured using the method described in section \ref{impact_of_temperature}.
As shown in Table \ref{table_diversity}, MoC significantly increases the diversity of the hypotheses. 
This indicates that our methodology effectively minimizes semantically redundant hypotheses, which leads to improvements in inductive reasoning performance.

\subsection{Analysis and Discussion}
\label{analysis_and_discussion}
In this section, we raise several research questions related to MoC and address them.

\paragraph{Does the benefit of MoC arise from hypothesis diversity or from an improvement in reasoning?}

\begin{table*}[t]
\small
\centering
\begin{tabular}{@{}llllll@{}}
\toprule
\textbf{Approach}        & List Fns    & MiniARC       & MBPP+       & Playgol-str   & Average              \\ \midrule
Baseline (\textit{K}=16) & 46          & 6.9           & 40          & 78            & 42.7                 \\
MoC (\textit{C}=16, \textit{S}=1) & 51          & 10.8          & \textbf{48} & 78.7          & 47.1 (+4.4)          \\
MoC (\textit{C}=8, \textit{S}=2)  & \textbf{52} & \textbf{12.3} & 46          & 80            & \textbf{47.6 (+4.9)} \\
MoC (\textit{C}=4, \textit{S}=4)  & 48          & 9.2           & 43          & \textbf{80.7} & 45.2 (+2.5)          \\ \midrule
Baseline (\textit{K}=32) & 46          & 9.2           & 45          & 81.3          & 45.4                 \\
MoC (\textit{C}=32, \textit{S}=1) & 51          & 13.1          & 47          & 80            & 47.8 (+2.4)          \\
MoC (\textit{C}=16, \textit{S}=2) & 54          & \textbf{13.8} & \textbf{55} & \textbf{85.3} & \textbf{52.0 (+6.6)} \\
MoC (\textit{C}=8, \textit{S}=4)  & \textbf{55} & 13.1          & 46          & 81.3          & 48.9 (+3.5)          \\ \bottomrule
\end{tabular}
\caption{Test accuracy (\%) of baseline and MoC with varying $C$ (number of concepts) and $S$ (number of hypotheses per concept). Base LLM is GPT-4o-mini. The numbers in parentheses indicate the improvement compared to the baseline.}
\label{table_multiple}
\end{table*}

It has been observed that LLMs can improve their reasoning ability by generating intermediate steps before the final answer, a method also known as Chain-of-Thought (CoT) prompting \citep{cot}. 
From this perspective, MoC, which first generates elementary concepts and then uses them to create hypotheses, shares similarities with such CoT prompting approaches.

To isolate the benefits of the reasoning ability from that on the hypothesis diversity, we set the temperature to 0 and generate only one hypothesis to observe if there is an improvement in accuracy. 
As shown in Table \ref{table_greedy}, the MoC method does not show significant gains compared to the greedy baseline.
Although there are certain model-dataset pairs where performance significantly improved (e.g., GPT-4o-mini on Playgol-str), no consistent trend was observed. 
Therefore, we conclude that the performance improvements observed with MoC in Table \ref{table_main} and \ref{table_scaling} are mostly due to the increase in hypothesis diversity rather than improvements in reasoning ability.

\paragraph{What impact does scaling compute have on the MoC?}
In the sampling-based inductive reasoning discussed in this work, if a larger compute budget is provided, performance improvement is almost guaranteed by sampling additional hypotheses. 
We investigate whether MoC shows a similar scaling effect, as observed in Figure \ref{figure_temperature_acc}, when the number of generated hypotheses $K$ is set higher.
We use GPT-4o-mini as base LLM for the scaling experiment.

\begin{figure}[H]
    \centering
    \includegraphics[width=0.7\columnwidth]{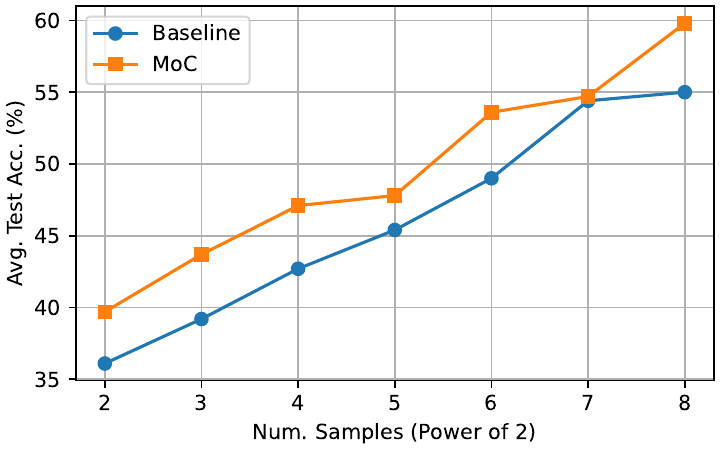}
    \caption{
    Test accuracy (\%) averaged over 4 datasets.
    }
    \label{figure_scaling}
    \vspace{-10pt}
\end{figure}

As shown in Figure \ref{figure_scaling}, both the IID baseline and MoC demonstrate higher average accuracy with larger number of samples $K$. 
Additionally, MoC achieves similar performance while generating only half the number of hypotheses relative to the baseline, demonstrating a significant improvement in the efficacy of inductive reasoning.
Full results of scaling experiment are in Appendix \ref{appendix_additional_results}.

\paragraph{What impact does generating multiple hypotheses per concept have on performance?}
In previous experiments, we generate only one hypothesis using a single concept as a hint. 
However, some concepts carry richer meanings than others, which may lead to the generation of multiple hypotheses. 
Additionally, if there is an error in the process of implementing the Python program, resampling may generate an error-free implementation.

In Table \ref{table_multiple}, we examine how the number of concepts ($C$) and the number of hypotheses sampled per concept ($S$) affect performance.
When we compare results while keeping the total number of hypotheses the same, we find that sampling about two hypotheses per concept performs better than other settings.
In conclusion, balancing $C$ and $S$ is critical for achieving optimal performance.

\paragraph{Does MoC assist in solving highly challenging problems beyond improving efficiency?}

\begin{table*}[t]
\centering
\small
\begin{tabular}{@{}m{3cm} m{3cm} m{9cm}@{}}
\toprule
\multicolumn{1}{c}{\textbf{Train Examples}} & \multicolumn{1}{c}{\textbf{Concepts}} & \multicolumn{1}{c}{\textbf{Natural Language Hypothesis}} \\ \midrule
\begin{tabular}[b]{@{}l@{}} \includegraphics[width=3cm]{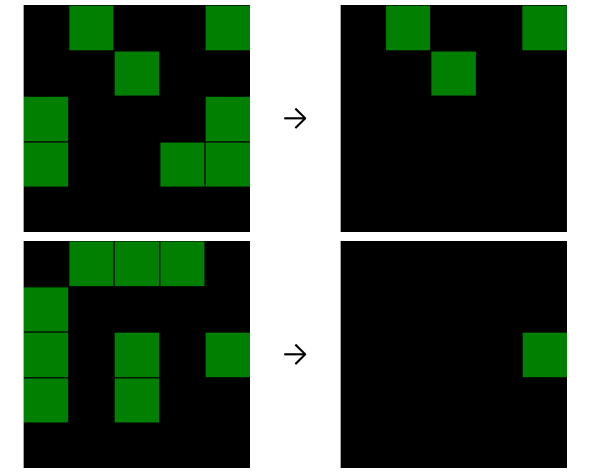} \\ ... \end{tabular} & \multicolumn{1}{c}{\begin{tabular}[B]{@{}c@{}} border, \\ corner, \\ \textbf{cell adjacency,} \\ identity transformation, \\ scan, ... \end{tabular}} & The transformation involves detecting continuous segments of \textbf{adjacent cells} containing the value '3' within the input grid. If any cell containing '3' has neighboring cells (up, down, left, or right) that also contain '3', the entire connected segment is cleared, meaning all cells in that segment are set to '0'. If a cell is isolated (has no adjacent '3's), it remains '3'. The process is repeated for the entire grid, adjusting the output based on this adjacency rule. \\ \midrule
\begin{tabular}[b]{@{}l@{}} $[2] \rightarrow$ False \\ $[10] \rightarrow$ True \\ $[61] \rightarrow$ False \\ $[55] \rightarrow$ True \\ ... \end{tabular} & \multicolumn{1}{c}{\begin{tabular}[B]{@{}c@{}} even numbers, \\ odd numbers, \\ \textbf{prime numbers,} \\ composite numbers, \\ divisibility, ... \end{tabular}} & The transformation checks if the input number is either a composite number or the number 1. If the number is \textbf{prime} (greater than 1 and only divisible by 1 and itself), the output is False. For all other inputs, including the number 1, the output is True. \\ \bottomrule
\end{tabular}
\caption{
Two challenging examples from MiniARC and MBPP+, where the IID baseline fails to generate a valid hypothesis from over 500 hypothesis samples.
MoC solves these problems correctly with only 64 hypotheses.
The concepts in \textbf{boldface} formulate the correct hypothesis.
}
\vspace{-10pt}
\label{table_example}
\end{table*}

The two examples in Table \ref{table_example} represent highly challenging problems in which more than 500 IID samples from GPT-4o-mini fail to identify the correct rule. 
In these problems, MoC effectively discovers the rule using far less compute ($K=64$).
This indicates that challenging problems with a low probability of being solved through IID sampling can be effectively resolved using the MoC approach.
In Appendix \ref{appendix_generated_concepts}, we provide a full list of generated concepts for these problems.

\section{Related Work}

\subsection{Automatic Inductive Reasoning}
Automating the process of inductive reasoning has long been a significant challenge in AI, due to its central role in achieving human-level intelligence \citep{humanlevel, ellis2023humanlike, dreamcoder} and its practical applications, such as Programming-by-Example (PBE) \citep{pmlr-v28-menon13, flashfill++, Gulwani}. 
Before the advent of LLMs, neural networks were widely used to guide program search but their search spaces are often limited by domain-specific languages (DSLs) \citep{executionguided, crossbeam, property_signature, bustle, clarke, exec_filter}.

Recently, LLMs have significantly transformed the paradigm of automatic inductive reasoning.
They are employed mainly in two ways: directly prompting with training input-output pairs to predict outputs for unseen inputs \citep{conceptarc, general_pattern, emergent_analogical, itd}, or formulating hypotheses in natural language or executable programs and applying them to unseen test inputs \citep{li2024programming, case2code, deer}.

Iterative refinement is a popular methodology in the latter domain \citep{hypothesis_search, phenomenal, greenblatt, code_repair}; however, they are sequential in nature and can slow down inference.
Furthermore, it has been suggested that LLMs lack the ability to self-correct \citep{huang2024large}, and that such refinement offers no substantial advantages compared to drawing more IID samples \citep{acquaviva2024overcoming, olausson2024is}.

\subsection{Reasoning in Inference Time of LLMs}
Recently, various methodologies have been explored to enhance the reasoning abilities of LLMs in inference time \citep{cot, zheng2024take}.
When allocated an increased inference budget, they are capable of solving complex reasoning tasks \citep{brown2024largelanguagemonkeysscaling, snell2024scaling}. 
By leveraging these properties, LLMs have shown promising reasoning capabilities to solving diverse reasoning tasks within a few iterative processes. 
For example, \citet{yao2023treethoughtsdeliberateproblem,lanchantin2023learningreasonmemorizeselfnotes} suggest an iterative generate-and-self-reflect approach to address complex reasoning tasks. 
\citet{kumar2024traininglanguagemodelsselfcorrect} show the self-correction capabilities of LLMs, and \citet{zhou-etal-2024-paraphrase} illustrate how paraphrasing input prompts can enhance reasoning capabilities.
In addition, many concurrent works have proposed methodologies to increase the diversity of responses to perform more efficient search \cite{wen2024synthesize, wang2024planning, light2024scattered, real}.

\section{Conclusion}
In this paper, we study the diversity of hypotheses generated by an LLM during inductive reasoning. 
We find that, even if incorrect, these hypotheses are predominantly redundant, leading to wasted compute. 
Increasing the LLM's temperature initially helps, but benefits saturate at certain point and excessively high temperatures cause text degeneration. 
Therefore, we propose a method called Mixture of Concepts (MoC); the core idea is to generate non-redundant concepts and using these to formulate hypotheses. 
Our approach significantly improves performance across various datasets and models.

\section*{Limitations}
Our MoC methodology requires additional computation during the concept proposal process prior to hypothesis generation. 
Additionally, during the hypothesis generation phase, each concept is used as a hint to construct prompts, so the model should encode more tokens compared to IID sampling. 
However, this incurs relatively small overhead compared to refinement or summarization-based methods found in existing works. 

Furthermore, the actual process of inductive reasoning in humans is far more complex than that of MoC. 
Humans are capable of solving difficult problems by transforming concepts and composing multiple concepts in various ways \cite{humanlevel}. 
These aspects have not been addressed in this study.

Lastly, it is not yet known how our approach affects the safety of LLMs as it may enhance the diversity of LLM responses in unexpected way. 
While such risks are minimal in the programming tasks we primarily dealt with, in the context of reasoning in natural language, there is a risk that increasing response diversity could amplify issues such as social bias and toxicity \cite{latte}.

\section*{Acknowledgments}
This work was partly supported by the Institute of Information \& Communications Technology Planning \& Evaluation(IITP)-ITRC(Information Technology Research Center) grant funded by the Korea government(MSIT)(IITP-2025-RS-2024-00437633, 50\%), Institute of Information \& communications Technology Planning \& Evaluation (IITP) grant funded by the Korea government(MSIT) [RS-2021-II211343, Artificial Intelligence Graduate School Program (Seoul National University) \& RS-2021-II212068, Artificial Intelligence Innovation Hub (Artificial Intelligence Institute, Seoul National University)], and the BK21 FOUR program of the Education and Research Program for Future ICT Pioneers, Seoul National University in 2024.
K. Jung is with ASRI, Seoul National University, Korea.

\bibliography{custom}

\clearpage
\appendix

\section{Prompts}
\label{appendix_prompt}
Here, we provide prompts for concept proposal (MoC) and hypothesis generation (baseline and MoC).
These prompts were constructed based on those from Hypothesis Search \citep{hypothesis_search}.

\begin{tcolorbox}[colback=white,colframe=black,title=Prompt for Hypothesis Generation: Baseline]
\lstset{
    basicstyle=\ttfamily\footnotesize,
    breaklines=true,
    frame=none,
    showstringspaces=false
}
\begin{lstlisting}
You will be given a list of input-output pairs. There is a SINGLE pattern that transforms each input to the corresponding output.
First, output a hypothesis for the transformation in natural language form.
Then, generate a Python function `fn` that maps the following inputs to their corresponding outputs.

Please format your hypothesis and Python function as follows:

```hypothesis
HYPOTHESIS
```

```python
def fn(x):
    # x is {INPUT_FORMAT}
    # Your code here
    return y # y is {OUTPUT_FORMAT}
```

Input-output pairs:
{TRAIN_EXAMPLES}
Hypothesis and Python function:

\end{lstlisting}
\end{tcolorbox}

\begin{tcolorbox}[colback=white,colframe=black,title=Prompt for Concept Proposal: MoC]
\lstset{
    basicstyle=\ttfamily\footnotesize,
    breaklines=true,
    frame=none,
    showstringspaces=false
}
\begin{lstlisting}
You will be given a list of input-output pairs. There is a SINGLE pattern that transforms each input to the corresponding output.
List {K} elementary concepts that may be useful to induce the transformation pattern.
Format your response in json format (dictionary whose keys are indices and values are elementary concepts).

Input-output pairs:
{TRAIN_EXAMPLES}

Elementary concepts:

\end{lstlisting}
\end{tcolorbox}

\begin{tcolorbox}[colback=white,colframe=black,title=Prompt for Concept Proposal: MoC (GPT-4o)]
\lstset{
    basicstyle=\ttfamily\footnotesize,
    breaklines=true,
    frame=none,
    showstringspaces=false
}
\begin{lstlisting}
You will be given a list of input-output pairs. There is a SINGLE pattern that transforms each input to the corresponding output.
List {K} elementary concepts that may be useful to induce the transformation pattern.
The concepts should be diverse, simple and concise.
Format your response in json format (dictionary whose keys are indices and values are elementary concepts).

Input-output pairs:
{TRAIN_EXAMPLES}

Elementary concepts:

\end{lstlisting}
\end{tcolorbox}

\begin{tcolorbox}[colback=white,colframe=black,title=Prompt for Hypothesis Generation: MoC]
\lstset{
    basicstyle=\ttfamily\footnotesize,
    breaklines=true,
    frame=none,
    showstringspaces=false
}
\begin{lstlisting}
You will be given a list of input-output pairs. There is a SINGLE pattern that transforms each input to the corresponding output.
First, output a hypothesis for the transformation in natural language form. Use hint: {HINT}.
Then, generate a Python function `fn` that maps the following inputs to their corresponding outputs.

Please format your hypothesis and Python function as follows:

```hypothesis
HYPOTHESIS
```

```python
def fn(x):
    # x is {INPUT_FORMAT}
    # Your code here
    return y # y is {OUTPUT_FORMAT}
```

Input-output pairs:
{TRAIN_EXAMPLES}
Hypothesis and Python function:

\end{lstlisting}
\end{tcolorbox}

\begin{table*}[t]
\small
\centering
\begin{tabular}{@{}llllllll@{}}
\toprule
\textbf{Model}                 & $K$                 &          & List Fns & MiniARC & MBPP+ & Playgol-str & Average              \\ \midrule
\multirow{4}{*}{GPT-4o}        & \multirow{2}{*}{4}  & Baseline & 46       & 8.5     & 50    & 78          & 45.6                 \\
                               &                     & MoC      & 52       & 10.8    & 48    & 80          & 47.7 (+2.1) \\ \cmidrule(l){2-8} 
                               & \multirow{2}{*}{16} & Baseline & 53       & 14.6    & 54    & 83.3        & 51.2                 \\
                               &                     & MoC      & 62       & 16.9    & 59    & 83.3        & 55.3 (+4.1) \\ \midrule
\multirow{4}{*}{Llama 3.1 70B} & \multirow{2}{*}{4}  & Baseline & 36       & 7.7     & 40    & 60.7        & 36.1                 \\
                               &                     & MoC      & 37       & 5.4     & 42    & 64          & 37.1 (+1.0) \\ \cmidrule(l){2-8} 
                               & \multirow{2}{*}{16} & Baseline & 44       & 10.8    & 54    & 65.3        & 43.5                 \\
                               &                     & MoC      & 52       & 8.5     & 52    & 74          & 46.6 (+3.1) \\ \midrule
\multirow{4}{*}{Qwen2.5 72B}   & \multirow{2}{*}{4}  & Baseline & 39       & 5.4     & 37    & 70.7        & 38.0                 \\
                               &                     & MoC      & 49       & 7.7     & 44    & 70          & 42.7 (+4.7) \\ \cmidrule(l){2-8} 
                               & \multirow{2}{*}{16} & Baseline & 48       & 11.5    & 46    & 80.7        & 46.6                 \\
                               &                     & MoC      & 51       & 14.6    & 51    & 79.3        & 49.0 (+2.4) \\ \bottomrule
\end{tabular}
\caption{Test accuracy (\%) with varying $K$. The numbers in parentheses indicate the improvement compared to the baseline.}
\label{table_additional}
\vspace{-6pt}
\end{table*}

\begin{table*}[t]
\small
\centering
\begin{tabular}{@{}lllllll@{}}
\toprule
$K$                  &                 & List Fns & MiniARC & MBPP+ & Playgol-str & Average \\ \midrule
\multirow{2}{*}{4}   & Baseline        & 34       & 4.6     & 33    & 72.7        & 36.1  \\
                     & MoC             & 43       & 7.7     & 36    & 72          & 39.7  \\ \midrule
\multirow{2}{*}{8}   & Baseline        & 43       & 5.4     & 33    & 75.3        & 39.2  \\
                     & MoC             & 49       & 8.5     & 40    & 77.3        & 43.7    \\ \midrule
\multirow{2}{*}{16}  & Baseline        & 46       & 6.9     & 40    & 78          & 42.7  \\
                     & MoC             & 51       & 10.8    & 48    & 78.7        & 47.1  \\ \midrule
\multirow{2}{*}{32}  & Baseline        & 46       & 9.2     & 45    & 81.3        & 45.4  \\
                     & MoC             & 51       & 13.1    & 47    & 80          & 47.8  \\ \midrule
\multirow{2}{*}{64}  & Baseline        & 53       & 13.1    & 46    & 84          & 49.0  \\
                     & MoC             & 57       & 15.4    & 58    & 84          & 53.6    \\ \midrule
\multirow{3}{*}{128} & Baseline        & 59       & 16.9    & 55    & 86.7        & 54.4    \\
                     & MoC             & 58       & 17.7    & 59    & 84          & 54.7  \\
                     & MoC (\textit{C}=64, \textit{S}=2) & 63       & 18.5    & 60    & 86.7        & 57.1   \\ \midrule
\multirow{3}{*}{256} & Baseline        & 62       & 16.2    & 55    & 86.7        & 55.0  \\
                     & MoC             & 64       & 20.8    & 69    & 85.3        & 59.8  \\
                     & MoC (\textit{C}=64, \textit{S}=4) & 64       & 20.8    & 66    & 84.7        & 58.9  \\ \bottomrule
\end{tabular}
\caption{Test accuracy (\%) with varying $K$ with GPT-4o-mini as a base LLM.}
\label{table_scaling}
\vspace{-8pt}
\end{table*}

In above prompts, \texttt{TRAIN\_EXAMPLES} represents all the train inputs and outputs of given inductive reasoning problem.
\texttt{INPUT\_FORMAT} and \texttt{OUTPUT\_FORMAT} specify input and output formats.
For instance, in List Functions dataset, both \texttt{INPUT\_FORMAT} and \texttt{OUTPUT\_FORMAT} are \texttt{List[int]}.
\texttt{K} represents the number of concepts $K$ in the main text.
\texttt{HINT} represents one of the concepts generated in the concept proposal stage.

We added the instruction ``The concepts should be diverse, simple and concise.'' in the concept proposal prompt for GPT-4o, because of excessively long and specific concepts.

\section{Experimental Details}
\label{appendix_experimental_details}
In this paper, we use 2 Nvidia A100 GPUs for running experiments with open-source LLMs.
The evaluation on four datasets takes approximately 1 to 2 days.
We conduct experiments using Llama 3.1\footnote{https://huggingface.co/meta-llama/Llama-3.1-70B-Instruct} and Qwen2.5\footnote{https://huggingface.co/Qwen/Qwen2.5-72B-Instruct} via Huggingface.

\section{Additional Results}
\label{appendix_additional_results}
Here, we provide additional experimental results.
Table \ref{table_additional} shows test accuracy of various models in various $K$.
In Table \ref{table_scaling}, we report full results of scaling experiment in section \ref{analysis_and_discussion}.

Also, we evaluate the performance of MoC when applied to GPT-4-0613 and compare it with the performance of Hypothesis Search on List Functions domain reported in their paper \cite{hypothesis_search}. 
According to Figure A.5 of the Hypothesis Search paper, the performance when generating 8 hypotheses and programs per problem appears to be 59 (although this is based on reading the graph). 
In comparison, the performance of MoC with $K=8$ using GPT-4-0613 was 62, indicating that MoC outperforms under similar conditions.

\newpage
\section{Generated Concepts}
\label{appendix_generated_concepts}
We provide a complete list of 64 concepts for the MiniARC and MBPP+ problems shown in Table \ref{table_example}.

\begin{tcolorbox}[colback=white,colframe=black,title=64 Concepts Proposed by MoC: MiniARC]
\lstset{
    basicstyle=\ttfamily\tiny,
    breaklines=true,
    frame=none,
    showstringspaces=false
}
\begin{lstlisting}
"matrix",
"grid",
"element",
"zero",
"non-zero",
"transpose",
"row",
"column",
"diagonal",
"sum",
"element removal",
"max",
"min",
"count",
"identity matrix",
"row operation",
"column operation",
"boolean logic",
"pattern recognition",
"transformation",
"filter",
"morhological operations",
"average",
"median",
"mode",
"conditions",
"local neighborhood",
"border",
"corner",
"cell adjacency",
"identity transformation",
"scan",
"order of operations",
"data structure",
"selectivity",
"placement",
"swap",
"binary logic",
"iteration",
"mappings",
"shift",
"constant pattern",
"symmetry",
"rotation",
"reflection",
"submatrix",
"duplicates",
"translational symmetry",
"identification",
"data comparison",
"sequence",
"position",
"inversion",
"overlay",
"region selection",
"objective spacing",
"dynamic structure",
"permutation",
"normalization",
"clustering",
"threshold",
"classification",
"consolidation",
"extraction"
\end{lstlisting}
\end{tcolorbox}

\newpage

\begin{tcolorbox}[colback=white,colframe=black,title=64 Concepts Proposed by MoC: MBPP+]
\lstset{
    basicstyle=\ttfamily\tiny,
    breaklines=true,
    frame=none,
    showstringspaces=false
}
\begin{lstlisting}
"even numbers",
"odd numbers",
"prime numbers",
"composite numbers",
"divisibility",
"factors",
"square numbers",
"cube numbers",
"binary representation",
"decimal values",
"number properties",
"natural numbers",
"integer numbers",
"numerical sequences",
"mathematical reasoning",
"truth values",
"falsehood",
"high values",
"low values",
"parity",
"logical operators",
"mathematical operations",
"set theory",
"constructive proof",
"algorithmic thinking",
"binary classification",
"discrete mathematics",
"logical expressions",
"conditions",
"case analysis",
"function evaluation",
"numerical properties",
"mathematical inductions",
"problem-solving techniques",
"classification systems",
"set membership",
"theoretical foundations",
"logic gates",
"boolean values",
"truth tables",
"coordinate systems",
"rounding numbers",
"fibonacci sequence",
"modular arithmetic",
"graph theory",
"probability theory",
"prime factorization",
"algebra",
"geometry",
"calculus",
"number theory",
"functional mathematics",
"mathematical logic",
"distributions",
"linear equations",
"asymptotic analysis",
"complex numbers",
"real numbers",
"irrational numbers",
"even-odd tests",
"theoretical number systems",
"order of magnitude",
"negation",
"quantitative reasoning"
\end{lstlisting}
\end{tcolorbox}

\section{Information about Use of AI Assistants}

In writing this paper, we utilized ChatGPT\footnote{https://chatgpt.com/} for paraphrasing.

\end{document}